\documentclass{jdmdh}
\usepackage{array}
\usepackage{pgfplots}
\usepackage{amsmath}
\usepackage{amsfonts}

\title{Variable selection for clustering with Gaussian mixture models: state of the art}
\author[1]{Abdelghafour Talibi}
\author[1]{Boujem\^aa Achchab}
\author[2]{Rafik Lasri}
\affil[1]{LAMSAD Laboratory, University of Hasan 1st, ESTB, Berrechid Morocco}
\affil[2]{SAAT Laboratory, University of Abdelmalek Essaâdi, FPL, Larache Morocco}

\corrauthor{Abdelghafour Talibi}{a.talibi@uhp.ac.ma}

\begin{document}

\maketitle

\abstract{The mixture models have become widely used in clustering, given its probabilistic framework in which it’s based, however, for modern databases that are characterized by their large size, these models behave disappointingly in setting out the model, making essential the selection of relevant variables for this type of clustering. After recalling the basics of clustering based on a model, this article will examine the variable selection methods for model-based clustering, as well as presenting opportunities for improvement of these methods.}


\section{Introduction}

\strut
\vspace{-4ex}

Clustering aims to classify objects of a population in groups, where the objects in the same group are similar to each other, and the objects in different groups are dissimilar. Unlike the supervised classification where the number of groups is known in advance, at least for a sample, in the case of clustering, it is unknown how many groups and it remains to be estimated. In fact, many fields of research used clustering methods on the data, in order to obtain groups that allow understanding and interpreting the phenomenon studied.
There is a very large family of clustering methods. One of these is called partitioning methods that are based on heuristics or geometric procedures defined by measure of proximity between observations in the same group, or between the observations in different groups, such as hierarchical clustering (\citet{ward}) and K-means (\citet{macqueen}) based on the minimization of the within-cluster sum of squares (WCSS) which is the distance between the observations belonging to the same cluster. Another family is based on a probabilistic framework called mixture model, in this family; the classification problem is approached by a probabilistic approach. This approach, as its name suggests, uses probabilistic modeling. The goal is always the same: to establish an automatic classification of individuals in homogeneous groups. Here the meaning given to the homogeneity of the groups is different: it is no longer based on geometric considerations but relies on the analysis of the probability distribution of the population. The Gaussian mixture models is the most commonly used model. The notion of homogeneity is reflected by the fact that the observations which are in the same group are coming from the same Gaussian distribution.\\

These mixing models presented by \citet{wolfe}, \citet{scott} and \citet{duda}, and then studied by \citet{MclaBas}, \citet{MclaBas}, \citet{Banf} or \citet{Fraley} and \citet{Fraraf}, have many advantages, such as their flexibility allowing to model a wide variety of random phenomena, due to their ability to adapt to many situations, and the possibility of being statistically interpreted, but, given the large number of variables present in modern databases, these model-based methods are over-parameterized in such a situation, resulting a poor performance. Moreover, this abundance of descriptive variables may seem an asset to determine a proper clustering of data. However, only a subset of these descriptive variables may be relevant for clustering, other variables may be redundant or even non-significant for the classification. In order to consider only the information required for the clustering, the selection of relevant variables must be considered, which will both improve the clustering process and facilitate the interpretation of the clustering results obtained.\\
Some studies have been focused on the selection variables in clustering; the main difficulty lies in the construction of a criterion to guide the selection of variables but cannot be based on the labels. The proposed methods are generally classified into filter or wrapper by the terminology borrowed from the supervised case, introduced by\citet{kohjohn}. Methods known filter treat the problem of variable selection regardless of the classification process. Among these methods, one can cite the work of \citet{dash} and \citet{jouve}. In contrast, the wrapper methods of variable selection procedures are included in the classification process.\\
This article is organized as follows. Section 2 briefly reviews the basics of clustering by mixture model and its estimation with the EM algorithm. Next, the section 3 respectively present the wrapper methods for selecting variables in the Bayesian approach (model selection) and under the frequentist approach (penalization of the likelihood). Finally, some methods are tested on real data set.

\section{The mixture model and the EM algorithm}

\subsection{The mixture model}
Mixture models have recently received much attention, given the intuitive idea that a population is composed of several groups as well as their flexibility to model a wide variety of phenomena.\\
The Gaussian mixture models are based on the idea that each group is represented by a multivariate Gaussian distribution, since each observation $x_i$  $i={1,..., n}$ is a vector $(x_{i1},..., x_{iJ} )$  with $x_{ij}$ the value of the variable $j$ for the observation $x_i$, with different parameters of the distributions of other groups, and the total population is represented by a mixture of these Gaussian distributions.
The general form of the likelihood of a mixture model with $K$ component for a single observation $x_i$:

\begin{equation}
  L(x_i;\theta)= \sum_{k=1}^{K}\pi_{k}f_{k}(x_i;\theta_{k})
\end{equation}

With,
$\pi_{1},..., \pi_{K}$: the mixture proportions,\\
$f_k$ : the Gaussian distribution of the component $k$,

\begin{equation}
  f_{k}(x_i;\theta_{k})= \frac{1}{(2\pi)^{J/2}|V_k|^{1/2}}\exp(-\frac{1}{2}(x-m_k)^{t}V_{k}^{-1}(x-m_k))
\end{equation}

$\theta_{k}=\{m_k, V_k\}$: the parameters of $f_k$  the Gaussian distribution of the component $k$,\\
$m_k$: the means vector of the component $k$,\\
$V_k$: the covariance matrix of the component $k$.\\

And the general form of the likelihood of a mixture model with $K$ components for $n$ observations $x_i$:

\begin{equation}
  L(x;\theta)= \prod_{i=1}^{n}[\sum_{k=1}^{K}\pi_{k}f_{k}(x_i;\theta_{k})]
\end{equation}

The parameter vector is then $\theta = (\pi_{1},..., \pi_{K}, m_1,..., m_K, V_1,..., V_K)$.\\

\subsection{The mixture model and clustering}
Clustering aims to classify the observations $x_i$, $i={1,..., n}$, representing a population composed of $n$ observations, where the observations $x_i$ are measured on $J$ variables $(x_{i1},...,x_{iJ} $), to  $K$ groups. let’s $(G_1,..., G_K)$ be a partition of the data. This data partition can be formalized by $z = (z_1,..., z_n)$, where the $n$ vectors $z_i = (z_{i1},..., z_{iK})$ are such that; $z_{ik}=1$ if the observation $x_i$ belong to the class $k$ and $z_{ik}=0$ otherwise.\\

The data clustering can be obtained by "mixture approach", which aims to estimate the parameters of the mixture and then classify the data by assigning each observation to the class with the highest membership probability. Or, by "classification approach", which consider $z$ also as a parameter to be estimated.

\subsubsection{The EM algorithm}

Even in the mixture approach, the maximum likelihood is difficult to solve because it represents a products, it is common to maximize the log-likelihood:

\begin{equation}
  \log L(x;\theta)= \sum_{i=1}^{n}\log[\sum_{k=1}^{K}\pi_{k}f_{k}(x_i;\theta_{k})]
\end{equation}

Maximizing the log-likelihood is even difficult, using the EM algorithm (Expectation Maximization algorithm) proposed by \citet{dempster}, and studied by \citet{mclakris}, is the most commonly used which is based on the completed log-likelihood:

\begin{equation}
  \log L_c(x;\theta)= \sum_{i=1}^{n}\sum_{k=1}^{K}z_{ik}[\log\pi_{k}+\log f_{k}(x_i;\theta_{k})]
\end{equation}

With :  $z_{ik}=1$ if the observation $x_i$ belong to the component $k$, otherwise $z_{ik}=0$.\\

The EM algorithm is based on maximizing successive iterations of the expectation of the complete penalized log-likelihood conditionally to the observations $x$ and a current value $\theta^{(r)}$ of the parameter vector,

\begin{equation}
  Q(\theta|\theta^{(r)})= E[\log L_c(x,z;\theta)|x, \theta^{(r)}]
\end{equation}

After the initialization of the vector of parameters $\theta^{(1)}$, this algorithm alternates between the following two steps. At the $r^{th}$ iteration,
\begin{itemize}
  \item E-step: This step consists to calculate the expectation $Q(\theta|\theta^{(r)})$, returning to express the conditional probabilities denoted $\hat{z}_{ik}^{(r)}$ that $x_i$ belongs to the component $k$:
\begin{equation}
  \hat{z}_{ik}^{(r)}=P(z_{ik}=1|x_i,\hat{\theta}^{(r)})
= \frac{\hat{\pi}_{k}^{(r)}f_{k}(x_i;\hat{m}_{k}^{(r)}, \hat{V}_{k}^{(r)})}{\sum_{k=1}^{K}\hat{\pi}_{k}^{(r)}f_{k}(x_i;\hat{m}_{k}^{(r)}, \hat{V}_{k}^{(r)})}
\end{equation}

  \item M-step: This maximization step consist to determine the vector of the parameters $\hat{\theta}^{(r+1)}$ maximizing $Q(\theta|\theta^{(r)})$.
This is equivalent to determine:\\
The proportions vector maximizing $Q(\theta|\theta^{(r)})$:
\begin{equation}
  \frac{\partial Q}{\partial \pi_{k}}= 0 \Rightarrow \hat{\pi}_{k}^{r+1} = \frac{\sum_{i=1}^{n}\hat{z}_{ik}^{(r)}}{n}
\end{equation}

The means vectors  maximizing $Q(\theta|\theta^{(r)})$:
\begin{equation}
  \frac{\partial Q}{\partial m_{k}}= 0 \Rightarrow \hat{m}_{k}^{r+1} = \frac{1}{\sum_{i=1}^{n}\hat{z}_{ik}^{(r)}}\sum_{i=1}^{n}\hat{z}_{ik}^{(r)}x_i
\end{equation}

The variance matrices maximizing $Q(\theta|\theta^{(r)})$:
\begin{equation}
  \frac{\partial Q}{\partial V_{k}}= 0 \Rightarrow \hat{V}_{k}^{r+1} = \frac{1}{\sum_{i=1}^{n}\hat{z}_{ik}^{(r)}}\sum_{i=1}^{n}\hat{z}_{ik}^{(r)}(x_i-\hat{m}_{k}^{r+1})(x_i-\hat{m}_{k}^{r+1})^{t}
\end{equation}

\end{itemize}

\subsubsection{Classification rule}

Once the estimation of the vector of parameters $\theta$ is done, we determine the best partition of the observations by assigning each individual to the class for which it has the highest probability of belonging.

\begin{equation}
  \hat{z}_{ik}=\frac{\hat{\pi}_{k}f_{k}(x_i;\hat{m}_{k}, \hat{V}_{k})}{\sum_{k=1}^{K}\hat{\pi}_{k}f_{k}(x_i;\hat{m}_{k}, \hat{V}_{k})}
\end{equation}

Each observation is finally assigned to the class for which the conditional probability is the largest.

\section{Variable selection for Gaussian mixture models}

Several recent studies were interested on variables selection for the clustering. The underlying idea of these works is that only a subset of all the existing variables is relevant for the clustering, the other variables are only harmful to the the clustering. The clustering task should be therefore made on the basis of the relevant variables, eliminating insignificant variables improved at one hand, the the clustering results, on the other hand, the interpretation of the resulting groups should be mitigated by the meaning of the selected variables. In clustering based on Gaussian mixture models, the selection of relevant variables was treated in two ways.\\
In the first way, authors such as \citet{law}, \citet{rafdean}, \citet{maugis}, \citet{maugisal} and \citet{maugismichel} treated the problem within a Bayesian framework, in fact, the selection of the relevant variables is restated here as a model selection problem with a determination of the role of each variable. In the second way, authors such as \citet{pan}, \citet{wang}, \citet{xie1} and \citet{xie2} dealt with the problem within a Frequentist framework, by introducing a penalty term in the likelihood function in order to select the relevant variables.

\subsection{Variable selection as a model selection problem}
\citet{law}, \citet{rafdean} and \citet{maugisal}, treat the variable selection problem for model-based clustering by determining the role of each variable. This determination of variables roles is recast in the work of \citet{law} as an estimation problem, while in the works of  \citet{rafdean} and \citet{maugisal}, it is treated as a model selection problem in the context of Gaussian mixture models, where they consider a parsimonious models based on a decomposition of the covariance matrix proposed by \citet{fralraft} and \citet{celeux}:
\begin{equation}
  V_k=\lambda_{k}D_{k}A_{k}D_{k}^{t}
\end{equation}

Where $\lambda_{k}$ is the largest eigenvalue of $V_k$ which controls the volume of the $k^{th}$ cluster, $D_{k}$ is the eigenvectors matrix  of $V_k$, which control the orientation of that cluster and $A_k$ is a diagonal matrix with the scaled eigenvalues as entries, which control the shape of that cluster. By imposing constraints on the various elements of this decomposition, a large range of models is available, ranging from the simple spherical models that have fixed shape to the least parsimonious model where all elements of the decomposition are allowed to vary across the clusters.

\citet{law} propose a solution to the variable selection problem in model-based clustering under the assumption that the irrelevant variables are independent of the relevant variables, by treating it as an estimation problem, which prevents any combinatorial search.
Instead of selecting a subset of variables, they estimate a set of actual values $\varphi_{j}$'s($\varphi_{j}\in[0, 1]$), with $\varphi_{j}=1$ if the variable $j$ is relevant for clustering ($\varphi_{j}=0$ otherwise) they define the quantities $\rho_{j}=P(\varphi_{j}=1)$, the probability that the variable $j$ is relevant, these quantities (one for each variable) that they call feature saliencies.

\begin{equation}
  p(x|\theta)=\sum_{k=1}^{K}\pi_{k}\prod_{j=1}^{J}(\rho_{j}f(x_{j}|\theta_{kj})+(1-\rho_{j})f(x_{j}|\zeta_{j}))
\end{equation}

Where $f(x_{j}|\theta_{kj})$ is the density function of the $j^th$ variable in the component $k$, and $f(x_{j}|\zeta_{j})$ a common density independent of the components.
Since they are in the presence of a model selection problem, it is necessary to avoid the situation where all saliencies take the maximum possible value. This is achieved by adopting the Minimum Message Length penalty(MML). The MML criterion encourages the saliencies of irrelevant variables to be equal to zero.

For \citet{rafdean}, the basic idea is to recast the variable selection problem as a comparison problem between competing models for all variables considered initially.
Comparing two nested subsets is equivalent to comparing two models, in one all variables that are in the largest subset are informative for the clustering, while in the other, the variables considered for exclusion are conditionally independent of the clustering given the variables included in both models. This comparison is performed using an approximation of Bayes Factors.

Contrary to \citet{law}, \citet{rafdean} do not consider that the irrelevant variables are independent with the relevant variables, but they define that all the irrelevant variables subset $Sc$ is dependent to all clustering variables subset $S$. the competing models are compared through Bayes factor of the log-likelihood by the BIC approximation. And the selected model maximizes the following quantity :
	
\begin{equation}
  (\hat{K},\hat{m},\hat{S})= \arg\max_{K,m,r,l,V}\{BIC_{clustering}(x^{S}|K,m)+BIC_{regression}(x^{S_{c}}|x^{S})\}
\end{equation}

Where $K$ is the number of components and $m\in M$ is a model that belongs to the family of parsimonious models available in the Software mclust (\citet{fralraft}).\\
The first term of (14) corresponds to the BIC approximation of Gaussian mixture model with $K$ components, the second, to the BIC approximation of a linear regression of the irrelevant variables in relation to the irrelevant variables.
They propose an algorithm, which every step, seeks to add the variable that improves the clustering as measured by BIC and evaluates if any of the current grouping variables can be eliminated. At each step, the best combination of number of components and clustering model is chosen. The algorithm stops when there is no improvement.

Indeed, as in the work of \citet{law}, the dependence of all the irrelevant variables to relevant ones seems questionable. To overcome the limits of the method of \citet{rafdean}, \citet{maugisal} consider firstly the subset $S$ which represents the relevant variables, and which includes a subset $R$ of the relevant variables related to a subset of irrelevant variables, and secondly, $Sc$ the complement of the subset $S$ which is divided into two subset: a subset $U$ of irrelevant variables which can be explained by linear regression to the subset $R$ and subset $W$ of irrelevant variables that is completely independent of all relevant variables, and try to find the subsets $F=(S,R,U,W)$, so their model is called $SRUW$. The selected model maximizes the following quantity:
	
\begin{equation}
\begin{split}
  (\hat{K},\hat{m},\hat{r},\hat{h},\hat{F})= \arg\max_{K,m,r,h,V}\{BIC_{clustering}(x^{S}|K,m)+ \\ BIC_{regression}(x^{U}|r,x^{R})\}+BIC_{ind}(x^{W}|l)\}
\quad
\end{split}
\end{equation}

The quantity (15) includes three terms, the first is the model-based clustering by a Gaussian mixture model with $K$ components on the subset $S$ and $m$ its shape chosen from a collection of 28 parsimonious models available in Mixmod software (\citet{bierna}), the second term represents a BIC approximation of the linear regression of the subset $U$ of irrelevant variables to the subset $R$ of relevant variables, $r$ is the form of the covariance matrix of the regression assumed to be spherical, diagonal or unconstrained. The last term corresponds to the BIC of a Gaussian distribution of the subset of irrelevant variables that are assumed to be independent of all relevant variables with $l$ the shape of its variance matrix assumed to be diagonal or spherical.

Also, \citet{maugismichel} present a new variable selection method for clustering. They re-form the variable selection problem of clustering as a model selection problem in the context of density estimation. They assume that the observed sample come from an unknown probability distribution with density $s$. A specific model collection is defined: Each model $S(K, v)$ corresponds to a particular clustering situation where $K$  is the number of cluster and v is the subset of relevant variables. A density $t$ belonging to $S(K, v)$ is decomposed in a density of a Gaussian mixture model with $K$ components on the subset $v$ of the relevant variables and a multidimensional Gaussian density on the other variables. The problem is reformulated as the choice of a model from a collection, as this choice automatically leads to a clustering of the data and a selection of variables. Thus, a data-driven criterion is necessary to select the "best" model from a model collection. This criterion depends on unknown multiplicative constant to be evaluated in practice. A heuristic method called "slope" is proposed and tested for this problem.
Their idea is that on the irrelevant variables, since the data are centered, individuals have a mean equal to zero, and these variables do not allow distinguishing different groups. So on these variables, the data is assumed to follow a common spherical Gaussian distribution with mean vector equal to zero.
While, on the relevant variables, the means vector of the different components are free and the data are assumed to have a completely free and positive-definite covariance matrix. On these variables the mixture model is selected from the family:

\begin{equation}
  L_{(K,\alpha)}=x\in \mathbb{R}^\alpha \rightarrow \sum_{k=1}^{K}\pi_{k}f(x;m_{k},V_{k})
\end{equation}

\begin{equation}
With  \forall k \in \{1,..., K\}, \pi_{k}\in]0,1[, \sum_{k=1}^{K}\pi_{k}=1, m_{k}\in[-a,a],(V_{1},..., V_{K})\in D_{K,\alpha}^{+}.
\end{equation}

Where $K$ is the number of components, $v$ is the index of the subset of relevant variables which their Cardinal is denoted $\alpha$, $a>0$ and $D_{K,\alpha}^{+}$ denotes a symmetric positive-definite matrix related to the specified form of Gaussian mixture.
On the irrelevant variables, a spherical Gaussian density belonging to the following family is considered:

\begin{equation}
  G_{(\alpha)}=x\in \mathbb{R}^{J-\alpha} \rightarrow f(x;0,\sigma^{2}I_{J-\alpha}), \sigma^{2}\in[\lambda_{m},\lambda_{M}]
\end{equation}

Thus, the Gaussian mixture family associated to the pairs $(K, v)$ is defined by:

\begin{equation}
 S_{(K, v)}= \{x\in \mathbb{R}^{J} \rightarrow f(x_{[v]})g(x_{[v^{c}]}); f\in L_{(K,\alpha)} g\in G_{(\alpha)}\}
\end{equation}

\subsection{Variable selection by likelihood penalization}

On the other hand, \citet{pan}, \citet{wang}, \citet{xie1}and \citet{xie2} select the relevant variables and perform clustering  by penalizing the log-likelihood function to maximize.
The penalized log-likelihood function has the following form:

\begin{equation}
  \log L_p(x;\theta)=\log L(x;\theta)-p_\lambda(\theta)
\end{equation}

Where $\log L(x_i;\theta)$ is the log-likelihood function and $p_\lambda(\theta)$ is the penalty function

The variable selection method of \citet{pan} is in the case of clustering for a small sample size and high dimension, when the data size exceeds the sample size.

Inspired by the penalized regression for selecting variables (Tibshirani, 1996; Fan and Li, 2001), they assume that the penalization can be as viable for variable selection in the context of model-based clustering and consequently they propose a clustering approach based on a penalized model. Specifically, the means $m_{k}$ specific to each cluster are adapted to a global mean $m$; with a penalty function appropriately selected, some variables means on the various components are estimated to be exactly $m$, allowing a selection of variables.
To facilitate the variable selection problem for "$J$ (Number variables) large, $n$ (sample size) small," \citet{pan} consider a diagonal covariance matrices common between all the clusters and they reduce and normalize the data so that each variable has a mean equal to 0 and a variance equal to 1. The form of their $l1$-norm penalty function is :

\begin{equation}
  p_\lambda(\theta)=\lambda\sum_{k=1}^{K}\sum_{j=1}^{J}|m_{kj}|
\end{equation}

Where $\lambda$ an hyper parameter which controls the level of desired sparsity and $m_{kj}$ the mean of the $j^th$ variable in the component $k$. Thus, given that the observations are normalized, if the means of a variable $j$ in each component are equal $m_{1j}=,...,=m_{Kj=0}$, this variable is considered to be irrelevant.
To select the value of the $K$ the number of components, and the value of the hyper parameter $\lambda$, they propose a modified BIC criterion.

\begin{equation}
  BIC=-2\log L_p(\hat{\theta})+\log(n)d
\end{equation}

Where $\hat{\theta}$ is the Maximum Likelihood Estimator (MLE) and $d = \dim(\theta)$ is the total number of unknown parameters (Fraley and Raftery, 1998).

Illuminated by the method of \citet{pan}, \citet{wang} also provides a method of variable selection for model- based clustering for a low sample size and large dimension, they consider a common diagonal covariance matrices between the clusters and they reduce and normalize the data so that each variable has a mean equal to 0 and a variance equal to 1.
Considering that the mean parameters in clusters associated with the same variable can be naturally grouped together, and intuitively should be treated as a group, they propose two new penalty functions, different from the penalty function of \citet{pan}, which does not take into consideration the "grouping" information in the data.
To eliminate non-informative variables, all $m_{kj}$, $k = 1,..., K$, should be equal to zero. However, the $l1$-norm penalty function proposed by Pan and Shen treated $m_{kj}$ individually, and, it does not use the information that $m_{kj}$ and $m_{k'j}$ are associated with the same variable $x_j$, and intuitively, they belong to a "group" and should be treated differently from $m_{kj'}$, which are associated with another variable $x_{j'}$. When the $j^{th}$ variable is uninformative, the $l1$-norm penalty function tends to shrink only a portion of $m_{kj}$, but not all to zero, where it fails to consider the $j^{th}$ variable as being irrelevant. This brought \citet{wang} to propose the $l\infty$-norm penalty function which shrinks, for each variable, the means in all clusters in order to identify the irrelevant variables:

\begin{equation}
  p_\lambda(\theta)=\lambda_{\infty}\sum_{j=1}^{J}\max_{k\in{1,..., K}}|m_{kj}|
\end{equation}

The $l\infty$-norm penalty function penalizes the maximum absolute value of $m_{kj}$, $k = 1,..., K$, if the maximum absolute for the $j^{th}$ variable is equal to zero, all other means for this variable in the different clusters are automatically reduced to be equal to zero.

For the same purpose as \citet{wang}, \citet{xie2} tried to overcome the limits of the $l1$-norm penalty function but in another way, they also found that the $l1$-norm penalty function treats the $m_{kj}$ individually, on the other side, a variable is irrelevant if $m_{1j}=,...,= m_{Kj} = 0$, in fact, to make a selection of relevant variables, it is natural to treat the means $m_{1j},..., m_{Kj}$ as a group of parameters and to construct a penalty encouraging all these means of a variable in the different clusters be equal to zero.
They observe that if considering the means in the clusters as a row vector, the direction of the regrouping $m_{1j},..., m_{Kj}$ is vertical and they call it the vertical means grouping (VMG), for which, they propose the following penalty :

\begin{equation}
  p_\lambda(\theta)=\lambda\sqrt{K}\sum_{j=1}^{J}\|m_{.j}\|
\end{equation}

Where $m_{.j} = (m_{1j}, m_{2j},..., m_{Kj})^{t}$ and $\|m_{.j}\| = \sqrt{\sum_{k=1}^{K}(m_{kj})^{2}}$ is the $l2$-norm penalty function on the means $m_{kj}$'s for $k = 1, 2,..., K$.\\
On the other hand, they consider also, that in some cases, through prior information, a group of variables is susceptible to be relevant or not, thus, they propose another group that considers this prior information, named the horizontal means grouping (HMG), in the case of common diagonal covariance matrices between the clusters. The grouping penalty proposed has the following :

\begin{equation}
  p_\lambda(\theta)=\lambda\sum_{k=1}^{K}\sum_{g=1}^{G}\sqrt{q_{g}}\|m_{k}^{g}\|
\end{equation}

Where $m_{k}^{m}$ corresponds to the mean of a group of variables, $\dim(m_{k}^{g}) = q_g$ and $\sum_{g=1}^{G}q_g=J$, with $J$ the number of variables.
A modified BIC is proposed as a model selection criterion to select the number of components $K$.

All these works already mentioned, assume that the clusters have a common diagonal covariance matrices, the common matrix implies that the clusters have the same size, which may be wrong in practice. Indeed, \citet{xie1} extend the method of \citet{pan}, by considering a cluster-specific diagonal covariance matrices, for which they have presented the following penalties functions:

\begin{equation}
 p_\lambda(\theta)=\lambda_1\sum_{k=1}^{K}\sum_{j=1}^{J}|m_{kj}|+\lambda_2\sum_{k=1}^{K}\sum_{j=1}^{J}|\sigma_{kj}^{2}-1|
\end{equation}

\begin{equation}
 p_\lambda(\theta)=\lambda_1\sum_{k=1}^{K}\sum_{j=1}^{J}|m_{kj}|+\lambda_2\sum_{k=1}^{K}\sum_{j=1}^{J}|\log\sigma_{kj}^{2}|
\end{equation}

The $l1$-norm penalty is used to force the irrelevant variables to have means $m_{kj}$ equal to 0, and a variances $\sigma_{kj}^{2}$ that are close to 1 to be exactly 1.

To insert the penalty, they propose a modified BIC as a model selection criterion.\\

Also, the diagonal covariance matrices assumption implies that the clusters have the same orientation, which may be also wrong in practice, incite \citet{zhupan}to propose a penalized likelihood approach for models with unconstrained covariance matrices. The first penalty proposed, allow the common covariance matrices to be unconstrained and have the following form:

\begin{equation}
 p_\lambda(\theta)=\lambda_1\sum_{k=1}^{K}\sum_{j=1}^{J}|m_{kj}|+\lambda_2\sum_{j=1}^{J}\sum_{l=1}^{J}|W_{jl}|
\end{equation}
Where $W_{kj}$ are the elements of $W=V^{-1}$ the inverse of the covariance matrix.\\
The second penalty function, allow also to the covariance to be different across the clusters, and have the following form:

\begin{equation}
 p_\lambda(\theta)=\lambda_1\sum_{k=1}^{K}\sum_{j=1}^{J}|m_{kj}|+\lambda_2\sum_{k=1}^{K}\sum_{j=1}^{J}\sum_{l=1}^{J}|W_{k,jl}|
\end{equation}
\subsection{Variable selection by combining likelihood penalization and model selection}

\citet{maynet} suggest the Lasso-MLE procedure, which combines both the method of \citet{pan} and the method of \citet{maugismichel}
The first step of their approach is to create a model sub-collection. As \citet{pan}, a $l1$-norm penalty is considered to get a sub-collection of models $\{S(K,J_r ),(K,J_r)\in M^l\}$, where $K$ is the number of components, and $J_r$ the subset of variables selected as relevant by the penalized maximum likelihood and $M^l$ is the penalized maximum likelihood. By changing each time $K$ the number of components of the mixture and the regularization parameter $\lambda$, an EM algorithm is used to maximize the penalized log-likelihood.

The second step consist to calculate the maximum likelihood $\hat{s}(K, Jr)$ for the sub-collections models $(K, Jr)$ obtained in the first step obtained by the penalized maximum likelihood, using the standard EM algorithm for each model . The third stage is devoted to the model selection problem, as in \citet{maugismichel}, a non asymptotic penalized criterion is proposed to solve the model selection problem.

\section{Numerical experiments}

In this section we test the method of \citet{rafdean} based on model selection and the method of \citet{wang} based on the on likelihood penalization on real data set.

The data set used in the experimentations is the data set IRIS (\citet{fisher}), which is, a reference and one of the most well known data sets in data mining, this data is composed of 150 observations of three plants of Iris (Iris setosa, Iris virginica and Iris versicolor), and measured on 4 continuous variables; sepal length(cm), sepal width (cm), petal length (cm) and petal width (cm).

The method of \citet{rafdean}, when applied with the true number of clusters $K=3$, select the model "$VEV$" and select the 3 variables sepal width, petal length and petal width as relevant for the clustering, and the variable sepal length. as being irrelevant for the clustering, as shown in Table ~\ref{tab:greedy}.

\begin{table}
  \newcolumntype{+}{>{\global\let\currentrowstyle\relax}}
  \newcolumntype{^}{>{\currentrowstyle}}
  \newcommand{\rowstyle}[1]{\gdef\currentrowstyle{#1}%
    #1\ignorespaces
  }

  \centering
  \begin{tabular}{+>{\bfseries}c^c^c^c^c^c}
    \hline
    \rowstyle{\bfseries}
    & Step & Variable proposed & Type of step & BIC difference & Decision\\
    & 1 & PL & Add & 167.549853 & Accepted\\
    & 2 & SW & Add & 52.954643 & Accepted\\
    & 3 & PW & Add & 26.366217 & Accepted\\
    & 4 & PW & Remove & 26.465480 & Rejected\\
    & 5 & SL & Add & 13.207518 & Accepted\\
    & 6 & SL & Remove & -4.393044 & Accepted\\
    \hline
  \end{tabular}

  \caption{Stepwise (forward/backward) results from the greedy search algorithm for the IRIS data set.}
  \label{tab:greedy}
\end{table}

While, the method of \citet{wang}, applied also with the true number of clusters, and with an hyper parameter $\lambda\in[1,12]$ select all the variables as being relevant for the clustering.
\section{Conclusion and discussion}

Model- based clustering has become a popular technique and a reference, but faced to a large data, this model suffers from the problem of dimensionality that over parameterize the model, to remedy this; many studies have been focused on the selection of variables to improve the clustering process and to facilitate the interpretation of the classification obtained. These works based on assumptions and restrictions may be further improved, and it is the aim of our research, whose main objective is to improve and/or propose new variable selection methods in this context, by a procedure that simultaneously selects the number of clusters and the relevant variables for the clustering. .

For example, the method of \citet{maugisal} require a considerable time to find the four subsets of variables, to minimize the required time, the use of a penalized likelihood approach, like the method of  \citet{xie2} based on the $l2$-norm  penalty or \citet{xie1} as a first step, can be tested to create sets of potentially relevant variables. Then, as a second step, use the method of \citet{maugisal} on these selected sets of potentially relevant variables.

\bibliographystyle{plainnat}

\end{document}